\documentclass[conference]{IEEEtran}
\IEEEoverridecommandlockouts
\usepackage{cite}
\usepackage{amsmath,amssymb,amsfonts}
\usepackage{graphicx}
\usepackage{textcomp}
\usepackage{xcolor}
\usepackage{multirow}
\usepackage[linesnumbered,ruled]{algorithm2e}
\usepackage[inkscapelatex=false]{svg}
\usepackage{subcaption}

\newcommand{\D}{{\mathcal{D}}}
\newcommand{\LL}{{\mathcal{L}}}
\newcommand{\EX}{{\mathbb E}}
\newcommand{\V}{{\mathcal{V}}}
\newcommand{\E}{{\mathcal{E}}}
\newcommand{\G}{{\mathcal{G}}}

\def\BibTeX{{\rm B\kern-.05em{\sc i\kern-.025em b}\kern-.08em
    T\kern-.1667em\lower.7ex\hbox{E}\kern-.125emX}}
\begin{document}

\title{Towards Cross-domain Few-shot Graph Anomaly Detection  \\ 
\thanks{$^{*}$Corresponding author}
}

\author{\IEEEauthorblockN{Anonymous Authors}}

\author{\IEEEauthorblockN{Jiazhen Chen$^{1}$, Sichao Fu$^{2, *}$, Zhibin Zhang${^{3}}$, Zheng Ma$^{4}$, Mingbin Feng$^{1,*}$, Tony S. Wirjanto$^{1}$, Qinmu Peng$^{2}$}

\IEEEauthorblockA{\textit{$^{1}$Department of Statistic and Actuarial Science, University of Waterloo}\\
\textit{$^{2}$School of Electronic Information and Communications, Huazhong University of Science and Technology}\\
\textit{$^{3}$Institute of Computing Technology, Chinese Academy of Sciences} \\
\textit{$^{4}$Cheriton School of Computer Science, University of Waterloo}\\
j385chen@uwaterloo.ca, fusichao$\_$upc@163.com, zhangzhibin@ict.ac.cn
\\ $\{$z43ma, ben.feng, twirjanto$\}$@uwaterloo.ca, pengqinmu@hust.edu.cn}}

\maketitle

\begin{abstract}

Few-shot graph anomaly detection (GAD) has recently garnered increasing attention, which aims to discern anomalous patterns among abundant unlabeled test nodes under the guidance of a limited number of labeled training nodes. Existing few-shot GAD approaches typically adopt meta-training methods trained on richly labeled auxiliary networks to facilitate rapid adaptation to target networks that possess sparse labels. However, these proposed methods often assume that the auxiliary and target networks exist in the same data distributions—an assumption rarely holds in practical settings. This paper explores a more prevalent and complex scenario of cross-domain few-shot GAD, where the goal is to identify anomalies within sparsely labeled target graphs using auxiliary graphs from a related, yet distinct domain. The challenge here is nontrivial owing to inherent data distribution discrepancies between the source and target domains, compounded by the uncertainties of sparse labeling in the target domain. In this paper, we propose a simple and effective framework, termed CDFS-GAD, specifically designed to tackle the aforementioned challenges. CDFS-GAD first introduces a domain-adaptive graph contrastive learning module, which is aimed at enhancing cross-domain feature alignment. Then, a prompt tuning module is further designed to extract domain-specific features tailored to each domain. Moreover, a domain-adaptive hypersphere classification loss is proposed to enhance the discrimination between normal and anomalous instances under minimal supervision, utilizing domain-sensitive norms. Lastly, a self-training strategy is introduced to further refine the predicted scores, enhancing its reliability in few-shot settings. Extensive experiments on twelve real-world cross-domain data pairs demonstrate the effectiveness of the proposed CDFS-GAD framework in comparison to various existing GAD methods including unsupervised, semi-supervised, few-shot and cross-domain GAD methods.

\end{abstract}


\section{Introduction}

Graph anomaly detection (GAD) is the process of identifying unusual nodes, edges or subgraphs within graph-structured data that significantly deviate from the majority. As a critical area in data mining, GAD finds extensive applications across diverse domains such as social networks, financial networks, and e-commerce~\cite{chen2022antibenford}~\cite{li2021relevance}~\cite{song2021subgraph}~\cite{fu2023towards}~\cite{fu2024finding}. These network structures often harbor various anomalies, which might manifest as spammers in social networks, money laundering in financial markets, or fraudulent transactions in e-commerce. The complexity of these datasets and the severe implications of overlooking anomalies have intensified the focus on GAD research in recent years. Among the various applications of GAD, detecting anomalous nodes is particularly crucial as it directly relates to pinpointing individual nodes exhibiting potentially harmful or irregular behaviors within networks. For example, in social networks, node-level GAD might involve detecting profiles that spread misinformation or engage in harmful activities. In financial networks, it could focus on identifying accounts involved in unusual transaction patterns that may indicate money laundering or financial fraud. In this paper, we mainly focus on addressing the intricate and practical challenges of node-level GAD.

Existing GAD techniques can be primarily categorized into unsupervised GAD, semi-supervised GAD, few-shot GAD and cross-domain GAD. Unsupervised GAD focuses on mining the intrinsic properties of graph structures and attributes solely based on unlabeled data. It operates under the assumption that anomalies will substantially deviate from learned patterns. Studies in this area predominantly employ reconstruction-based methods~\cite{ding2019deep}~\cite{ fan2020anomalydae}~\cite{luo2022comga} or contrastive learning-based approaches~\cite{liu2021anomaly}~\cite{zheng2021generative}~\cite{ zhang2022reconstruction}~\cite{duan2023graph}~\cite{pan2023prem} to identify these deviations. For instance, DOMINANT~\cite{ding2019deep} uses autoencoders to reconstruct graph structures and attributes, spotting anomalies through significant reconstruction errors in comparison to the majority of nodes in a network. PREM~\cite{pan2023prem} evaluates the cohesiveness of node embeddings with their neighborhoods using contrastive learning scores, based on the principle that anomalous nodes are substantially different from their surroundings. While unsupervised GAD is valuable in label-free scenarios, its effectiveness can be compromised by noise, leading to potential false positives or undetected anomalies.

Semi-supervised GAD aims to bridge the gap between unsupervised and fully supervised GAD by leveraging both labeled and unlabeled data to enhance anomaly detection performance. These methods typically utilize labeled examples to direct the identification of anomalies, while employing unlabeled data to improve model generalization~\cite{wang2019semi}~\cite{kumagai2021semi}~\cite{meng2021semi}~\cite{tian2023sad}. For instance, SemiGNN~\cite{wang2019semi} trains a binary classifier on labeled samples and enhances the learning of normal representations through a hierarchical attention mechanism. SemiADC~\cite{meng2021semi} uses generative adversarial networks to model the distribution of normal node features from unlabeled data, subsequently employing a binary classifier to differentiate these normal samples from labeled anomalies. SAD~\cite{tian2023sad} introduces a memory-enhanced deviation network to accentuate the differences between abnormal and normal nodes in dynamic graphs, supplemented by a contrastive learning module to leverage unlabeled data. Despite its potential advantages, semi-supervised GAD often hinges on the availability of an adequate number of labeled samples, a condition that is rarely met in real-world scenarios, thus hampering its effectiveness.

In recent years, few-shot GAD has garnered increasing attention for its ability to effectively operate in label-scarce environments. This line of research aims to detect anomalies using only a limited number of labeled anomalies~\cite{ding2021few}~\cite{ guo2022learning}~\cite{zheng2022unsupervised}~\cite{xu2023few}. For instance, ANEMONE-FS~\cite{jin2021anemone} employs a multi-scale contrastive learning framework that utilizes few-shot labeled anomalies to enhance model discrimination between normal and anomalous behaviors. Meta-GDN~\cite{ding2021few} introduces a graph deviation network, which is trained under a cross-network meta-learning framework. The framework involves meta-training across multiple auxiliary networks to establish a generalized model, subsequently fine-tuned on the target network with a few labeled examples. LHML~\cite{guo2022learning} introduces a dynamic learnable hypersphere that adjusts its decision boundary to include normal samples. It also applies cross-subgraph meta-learning for rapid adaptation from auxiliary to target networks. Despite the promising results of these methodologies, most of them rely on sufficiently labeled auxiliary networks that are presumed to share the same distributions with the target network. Nevertheless, acquiring a sufficient volume of labeled data in the same domain can be prohibitively costly and challenging in practical scenarios.

While acquiring sufficient labels within a single domain can be challenging in real-world scenarios, obtaining adequate labeled data from pre-existing, related domains is often feasible. For example, hotel and dining services from Yelp may share similar fraudulent review patterns due to overlapping user bases and behaviors. A few recent studies focus on tackling the cross-domain GAD problem, which leverages labeled data from a source domain to improve anomaly detection in an unlabeled target domain. For instance, COMMANDER~\cite{ding2021cross} adopts a graph-attentive encoder and domain discriminator to learn domain-invariant features. It further includes an anomaly classifier and an attribute decoder to detect anomalies that are both common and unique across domains. ACT~\cite{wang2023cross} employs contrastive learning to develop normal node representations in the target graph, and aligns these with source graph representations trained under sufficient source domain supervision. The framework is further enhanced by a self-training mechanism to refine detection accuracy. However, the efficacy of cross-domain GAD can be compromised without direct supervision from the target domain, especially when there are substantial cross-domain gaps.

In practice, it is often feasible to acquire a small number of labeled anomalies at a relatively low cost. Therefore, a more practical approach for the above-mentioned challenges is to leverage these scarce but valuable labeled anomalies from the target domain to guide the domain adaptation process. This aids in the better extraction and integration of both domain-invariant and domain-specific knowledge. In this study, we mainly focus on the cross-domain few-shot GAD problem. Specifically, we seek to improve GAD performance under label-scarce conditions with the help of adequate labeled data from a different but related domain. This task is complex owing to large domain shifts, which manifest as variances in graph structures, attributes, and anomaly distributions between the source and target domains~\cite{zhu2024graphcontrol}~\cite{ding2021cross}. These differences necessitate a robust model capable of assimilating common knowledge across diverse domains while preserving the unique characteristics inherent to each domain. Furthermore, the scarcity of labels in the target domain exacerbates the training complexity, introducing additional noise and uncertainty. Therefore, it is crucial to explore effective methodologies that can successfully transfer knowledge from the source domain with minimal supervision.

In this paper, a simple and effective framework is proposed to address the challenges of domain shift and label scarcity encountered in cross-domain few-shot GAD settings, termed CDFS-GAD. Specifically, CDFS-GAD introduces a domain-adaptive graph contrastive learning module, aiming to enhance intra-domain node representations while aligning cross-domain distributions via inter-domain graph contrastion. To further retain the unique characteristics of anomaly distribution in the target domain, a domain-specific prompt-tuning module is proposed that adapts the backbone encoder to the nuances of each domain without compromising domain-invariant feature extraction. To more effectively leverage the scarce labels of target domains, a domain-adaptive hypersphere classification loss is introduced. It encloses normal instances within a hypersphere, with the center adjusted contextually based on domain-invariant and domain-specific information. Post-training, a simple but effective self-training phase is introduced to further enhance the confidence of the model's prediction in few-shot settings.
\begin{figure*}[t]
    \centering
    \includegraphics[width=1\linewidth]{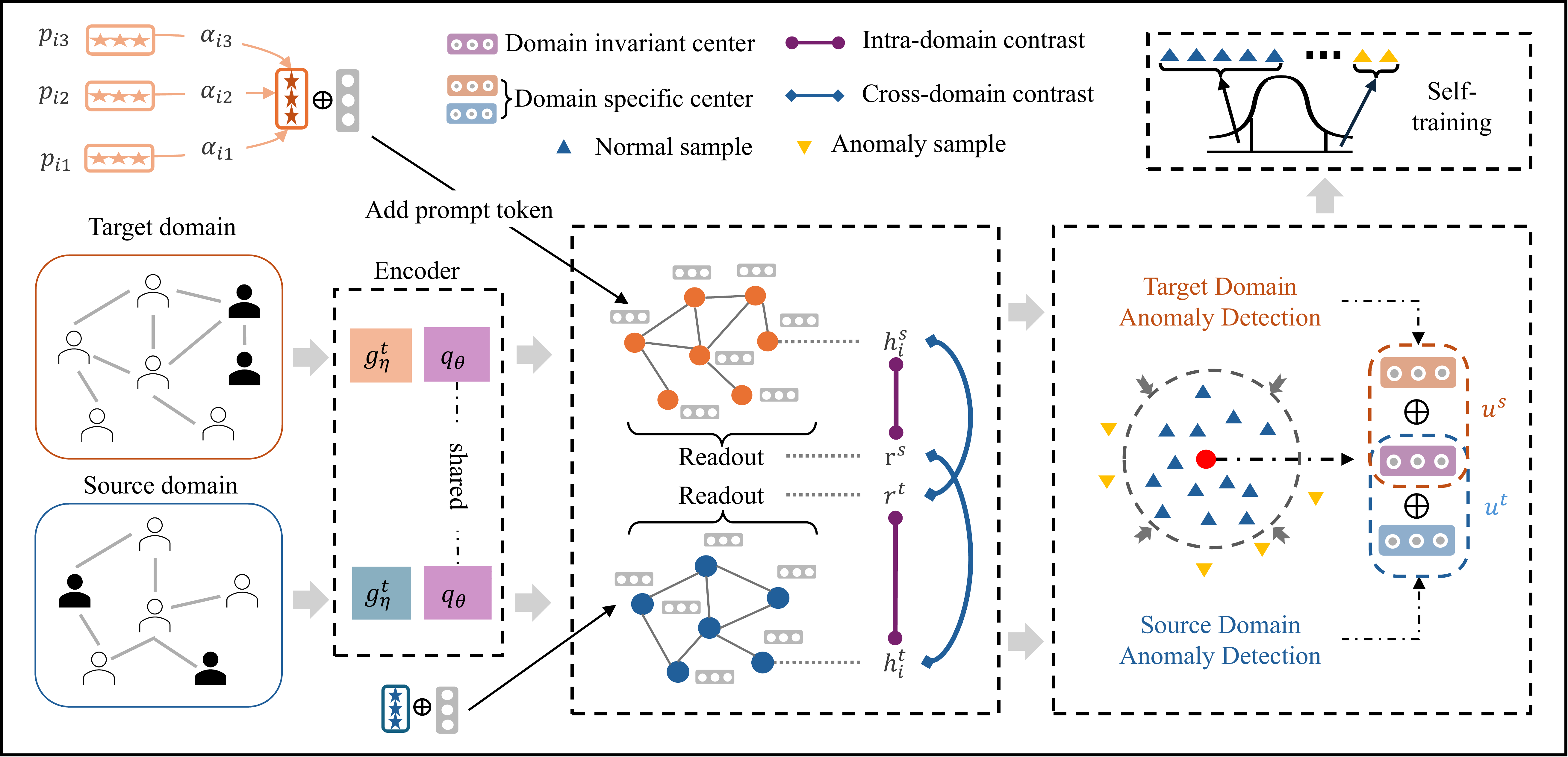}
    \caption{A diagram illustrating the proposed CDFS-GAD framework.}
    \label{fig: framework}
\end{figure*}

To summarize, our contributions are listed as follows:

\begin{itemize}

\item \textbf{Novel Problem}: To the best of our knowledge, we are the first to investigate and address the graph anomaly detection problem under the cross-domain few-shot setting.
    
\item \textbf{Enhanced Cross-domain Feature Extraction}: We propose a domain-adaptive graph contrastive learning module alongside a domain-specific prompt-tuning module. These components are designed to extract both common and unique features across two domains, optimizing for domain generalization.

\item \textbf{Domain-aware Few-shot Graph Anomaly Detection}: A domain-adaptive hypersphere classification loss is introduced, which shapes the detection boundary using domain insights under minimal supervision. A self-training process is implemented to further reinforce robust learning under few-shot conditions.

\item \textbf{Extensive Experimental Validation}: We conduct comprehensive experiments on twelve cross-domain pairs under varying few-shot conditions. The experiments demonstrate the superior performance of our method compared to existing GAD methods.
    
\end{itemize}

\section{Method}

\subsection{Problem Formulation}

In this paper, we study the cross-domain few-shot GAD problem on attributed graphs. An attributed graph is represented as $\G=(\V, \E, X)$, where $\V$ represents the set of nodes $\{v_1, v_2, ..., v_n\}$, $\E$ represents the set of edges $\{e_1, e_2, ..., e_m\}$, and $X = [x_1;...;x_n] \in \mathbb{R}^{n\times d}$ denotes the node attributes. Additionally, $A \in \mathbb{R}^{n \times n}$ denotes the adjacency matrix of $\mathcal{G}$, which encodes the edge connectivity structure of the graph. In the cross-domain few-shot GAD setting, we are given an attributed graph $\G^s=(\V^s, \E^s, X^s)$ from the source domain, and an attributed graph $\G^t=(\V^t, \E^t, X^t)$ from the target domain. For the source domain, we have access to the true label of each of the nodes in $\G^s$. Conversely, in the target domain, we only have access to a limited subset of labels ($|\V^t_l| \ll |\V^t_u|$), where $\V^t_l$ and $\V^t_u$ respectively denote the labeled anomalies and the unlabeled nodes in the target domain. 

The primary objective is to utilize the labeled data across both domains to identify anomalous nodes in $\G^t$. We aim to construct an anomaly detection model that calculates an anomaly score for each node in $\G^t$. The model will prioritize higher scores for anomalous nodes and lower scores for normal nodes, in alignment with common practices that frame GAD as a ranking problem~\cite{ding2019deep}.

\subsection{Overview}

The overall pipeline of CDFS-GAD is illustrated in Figure 1. CDFS-GAD initiates by jointly training a model using labeled data from both source and target domains (Section~\ref{sec: encoder}). To bridge the domain gap while enhancing the feature representation of normality patterns within each domain, we propose a domain-adaptive contrastive learning module as elaborated in Section~\ref{sec: contrasting learning}. This module includes both inter-graph and intra-graph contrasting components to align cross-domain features and refine within-domain normality representations. To further enhance the capability of the proposed CDFS-GAD to distinguish between domains, we incorporate unique, learnable prompt tokens into the backbone model. This integration improves the flexibility of the model to extract domain-specific features (Section~\ref{sec: prompt learning}). For model optimization, we introduce a domain-adaptive hypersphere classification loss, which effectively segregates normal instances from anomalies into a domain-sensitive hypersphere, as described in Section~\ref{sec: hsc loss}. Lastly, we introduce a self-training strategy that refines target domain predictions using pseudo-labels generated by the trained model, as discussed in Section~\ref{sec: self-training}.

\subsection{Feature Learning and Adaptation}

\subsubsection{Encoder}
\label{sec: encoder}

To address the challenge of limited label availability in the target domain, we commence by training a joint model based on labeled data from both the source domain and the target domain. We utilize GraphSAGE~\cite{hamilton2017inductive} as the backbone GNN encoder, which is shared across both domains. Owning to the inherent differences in node feature dimensionality and distribution between the two domains, we employ distinct MLP networks to preprocess the node features for each domain separately. These initial representations, tailored to their respective domains, are then inputted into the GNN encoder. In the following section, we denote the GNN encoder function as $q_{\theta}(\cdot)$, and the MLP mapping function of source and target domain to be $g^s_{\eta}(\cdot)$ and $g^t_{\xi}(\cdot)$, respectively.

\subsubsection{Domain-Adaptive Contrastive Learning}
\label{sec: contrasting learning}

Reliance solely on labeled data, particularly the scarce labels available in the target domain, may compromise the model's ability to generalize effectively. This may pose challenges in learning domain-invariant features from the source domain that are essential for the target domain. To address this, we introduce a domain-adaptive contrastive learning module that leverages the abundant unlabeled datasets available across both domains. This module is devised with two objectives: firstly, to improve the representation quality of normal nodes that constitute the majority of the dataset, and secondly, to align node representations between domains, thereby bridging the domain discrepancies.

Drawing inspiration from Deep Graph Infomax (DGI)~\cite{velickovic2019deep}, our proposed module consists of two distinct components: intra-domain and inter-domain contrastive learning. The intra-domain contrastive learning enhances within-domain representation by contrasting individual node representations against their corresponding graph-level representations. For node $i$ in a target domain, we define the positive pair as the node representation $h^t_i=q_{\theta}(g^t_{\xi}(x^t_i), \tilde{A^t})$ and the graph representation $r^t=\mathcal{R}(\{h^t_i|v_i \in \V^t\})$. Here, $\tilde{A}^t=A^t \odot M$ is the adjacency matrix after an edge-dropping operation, with $M$ being a binary mask created by sampling from a Bernoulli distribution, i.e., $m_{ij} \sim Bernoulli(p)$. The readout function $\mathcal{R}(\cdot)$ calculates the graph embedding by aggregating the node embeddings. Negative pairs are constructed using corrupted node representations $\tilde{h}^t_i = q_{\theta}(g^t_{\xi}(\tilde{x}^t_i), \tilde{A^t})$, where $\tilde{x}_i^t$ is generated by shuffling node features across the graph. The intra-domain contrastive loss for the target domain is then defined as:

\begin{small}
\begin{align}
    \LL^{t}_{intra} =  \EX_{v_i\sim \V^{t}} \left[ -\log{D(h^{t}_i, r^{t})} - \log (1-D(\Tilde{h}^{t}_i, r^{t})) \right]
\end{align}
\end{small}

where $D(\cdot, \cdot)$, defined as $D(x, y)=\sigma(x^tWy)$, computes the similarity between two representations.

On the other hand, the inter-domain contrastive learning module aims at aligning the feature distribution across domains. Recognizing the inherent challenge of establishing direct node mappings between source and target domains, we opt to contrast the overall graph representations of each domain.  Specifically, positive pairs are formed between the source and target graph representations, while negative pairs involve contrasting the corrupted source domain graph representation with the target domain representation. The resulting inter-domain contrastive loss is given by:
\begin{align}
    \LL^{t}_{inter} =   -\log{D(r^{t}, r^{s})} - \log (1-D(r^{t}, \Tilde{r}^{s})) 
\end{align}
where $\Tilde{r}^s$ represents the corrupted source domain graph representation, calculated by $\Tilde{r}^s=\mathcal{R}(\{\tilde{h}^s_i|v_i \in \V^s\})$.

Similarly, the inter and intra-domain contrastive learning is applied to the source domain, to ensure that the learning process is balanced across both domains. The final domain-adaptive contrastive loss is defined as:    
\begin{align}
    \LL_{contra} = \LL^{s}_{intra}+\LL^{s}_{inter} +\LL^{t}_{intra}+\LL^{t}_{inter}
    \label{eq: loss contra}
\end{align}

\subsubsection{Domain-Specific Prompt Learning}
\label{sec: prompt learning}

While the contrastive learning module and the shared encoder can facilitate the learning of domain-invariant features, it may inadvertently overlook the unique information intrinsic to each domain. To address this limitation, we augment the backbone encoder with a domain-specific prompt learning module. This enhancement is specifically designed to capture the distinct characteristics of each domain, while ensuring that it does not impede the ability of the encoder to learn domain-invariant features.

\SetKwComment{Comment}{/* }{ */}

\begin{algorithm}[t]

\caption{Training Process of CDFS-GAD}

\DontPrintSemicolon

\KwIn{$\G^s$ and $\G^t$, $N_a$ and $N_b$ as the epochs for joint training and self-training \;}

\KwOut{Anomaly scores for all nodes in $\G^t$\;}

\For{$i \leq N_a$}{  
  Obtain the augmented adjacency matrix $\tilde{A^t}$ and $\tilde{A^s}$\;
  
  Obtain the node representations $H^t, H^s \leftarrow q_{\theta}(g^t_{\xi}(X^t, \tilde{A^t})), q_{\theta}(g^s_{\eta}(X^s, \tilde{A^s}))$\;
  
  Obtain the graph representations $r^t, r^s \leftarrow \mathcal{R}(H^t), \mathcal{R}(H^s)$\;
  
  Compute the domain-adaptive contrastive loss  $\mathcal{L}_\text{contra}$ based on $H^t$, $H^s$, $r^t$, $r^s$ by Eq.~\ref{eq: loss contra}\;
  

    Obtain the node representations with prompt tokens $Z^t$, $Z^s$ by Eq.~\ref{eq: prompt token}\;
  
  Compute the domain-adaptive hypersphere classification losses $\mathcal{L}^s_\text{dahsc}$ and $\mathcal{L}^t_\text{dahsc}$ based on $Z^s$, $Z^t$ by Eq.~\ref{eq: loss dahsc}\;
  
  Take the gradient steps and update the parameters based on the loss $\mathcal{L}^s_{\text{dahsc}}+\mathcal{L}^t_{\text{dahsc}}+\alpha \mathcal{L}_{\text{contra}}$\;
}

Obtain pseudo-labels of target graph by Eq.~\ref{eq: pseudo label generation 1} and Eq.~\ref{eq: pseudo label generation 2}\;

\For{$i \leq N_b$}{
  obtain the node representations $H^t \leftarrow q_{\theta}(g^t_{\xi}(A^t, X^t))$\;
  
  Compute the loss $\mathcal{L}^t_\text{dahsc}$ for target domain\;
  
  Take gradients and update parameters\;
}

Compute anomaly score for the nodes in $\G^t$\;

\end{algorithm}

Prompt learning is a key technique in NLP that adapts pre-trained large language models to diverse tasks by transforming inputs into specific prompts~\cite{liu2023pre}. The idea has been adapted to graph-structured data, empowering pre-trained models to address a multitude of graph-related tasks via context-relevant prompts~\cite{fang2024universal}~\cite{sun2023all}~\cite{tan2023virtual}~\cite{liu2023graphprompt}. Motivated by recent studies, we incorporate prompt learning into our backbone model. This ensures that it complements, rather than compromises the extraction of domain-invariant features by the encoder.

We primarily follow the methodology proposed in~\cite{fang2024universal}, where each node is assigned a unique learnable prompt token. The prompt token is constructed as a weighted combination of a set of independent basis vectors $\{p_j\}_{j=1:m}$ with the weights calculated as:
\begin{align}
    \alpha_{ij}^{(l)} = \frac{\exp(h^{(l)}_i \cdot p_j)}{\sum_{k=1}^m{\exp(h^{(l)}_i \cdot p_k)}}
\end{align}
where $h^{(l)}_i$ represents the $l^{th}$-layer outputs of the GNN encoder for node $i$. Note that each domain has a set of different basis vectors, i.e., $\{p^s_j\}_{j=1:m}$ and $\{p^t_j\}_{j=1:m}$ for the source and target domains, respectively. Therefore, the prompt tokens can focus on learning the unique characteristics of each domain. 

Finally, the prompt-enhanced feature for each node of the $l^{th}$ layer is formulated by: 
\begin{align}
    z^{(l)}_i &= h^{(l)}_i + \sum_{j=1}^m \alpha^{(l)}_{ij} p_j \ 
    \label{eq: prompt token}
\end{align}
which allows $z^{(l)}_i$ to maintain its original information while being augmented with domain-specific insights encoded in the domain-specific prompt tokens. Then, $z^{(l)}_i$ is used as the input to the $(l+1)^{th}$ GNN layer.

\subsection{Model Training and Testing}

\subsubsection{Domain Adaptive Graph Anomaly Detection}
\label{sec: hsc loss}

We adopt the Hypersphere Classification (HSC) Loss~\cite{ruff2020rethinking}, which is tailored for anomaly detection in scenarios with scarce anomaly labels. The core idea of this loss function is to cluster normal samples around a central point while ensuring that anomalous samples are kept at a distance.  However, applying a single center for training across source and target domains may lack flexibility due to their distinct anomaly distributions. To circumvent this, we append the center with a domain-specific prompt for each domain. In this way, the loss function can dynamically adjust the center to accommodate the unique characteristics of each domain while leveraging commonalities.

Specifically, we define the center for the target domain as $c^{t}=u^{t}+c$, where $u^t$ denotes the target domain-specific prompt, and $c$ embodies the domain-invariant center common to both domains. Both $c$ and $u^t$ are learnable vectors. Then, we define the domain-adaptive hypersphere classification loss for the target domain as:

\begin{footnotesize}
    \begin{align}
        \LL^t_{dahsc} =\EX_{v_i\sim \G^{t}} \left[ -y^t_i \log{l(z^t_i, c^t)} - (1-y^t_i) \log{(1-l(z^t_i, c^t))} \right]
        \label{eq: loss dahsc}
    \end{align}
\end{footnotesize}
where $z^t_i$ and $y^t_i$ are the hidden representations and label for the node $i$ respectively, and $l(x, y)=\exp{(-\|x-y\|^2)}$. Similarly, the loss $\LL^{s}_{dahsc}$ is defined for the source domain, with the learnable center as $c^s=u^s + c$.

By integrating with the contrastive loss in the previous section, the overall training loss is defined as:
\begin{align}
    \LL = \LL^t_{dahsc} + \LL^s_{dahsc} +\alpha \LL_{contra}
\label{eq: loss total}
\end{align}
where $\alpha$ is a hyper-parameter controlling the contribution of the contrastive loss. During inference, the anomalous score of node $i$ for the target domain is obtained by:
\begin{align}
    s^t_i = 1-l(z_i^t, c^t)
    \label{eq: anomaly score}
\end{align}

\subsubsection{Self-Training}
\label{sec: self-training}

To mitigate the uncertainty inherent in the limited availability of labeled data within the target domain, we implement a simple but effective self-training strategy. This strategy creates a more reliable dataset for subsequent model refinement phases by designating pseudo-anomalous and pseudo-normal samples. We introduce percentile-based criteria for pseudo-label selection. Specifically, the pseudo-anomalous set $\D^{+}$ and the pseudo-normal set $\D^{-}$ are obtained as:
\begin{align}
\D^{+} &= \{v_i|s_i \text{ is in the top } \beta_1 \text{ percentile of scores}\} \label{eq: pseudo label generation 1} \\
\D^{-} &= \{v_i|s_i \text{ is in the bottom } \beta_2 \text{ percentile of scores}\}
\label{eq: pseudo label generation 2}
\end{align}
where $\beta_1$ and $\beta_2$ are hyperparameters representing the percentage of nodes to be included in the pseudo-anomalous and pseudo-normal sample sets, respectively. These pseudo-labels are then used to retrain the model using the hypersphere classification loss. The final anomaly score for each node is computed according to Eq.~\ref{eq: anomaly score}. The overall training procedure of CDFS-GAD is summarized in Algorithm 1.

\section{Experiments}

\subsection{Datasets}

In this study, the datasets utilized in the experiments contain YelpHotel, YelpRes, YelpNYC, and Amazon. The summary statistics of these four datasets are presented in Table~\ref{table: data statistics}. Specifically, YelpHotel comprises user reviews centered on the accommodation sector in Chicago, with nodes representing users and edges representing mutual hotel reviews~\cite{rayana2015collective}. Yelp employs a sophisticated filtering algorithm to categorize users as either regular or anomalous, according to their review activities. YelpRes follows a similar structure, which focuses on reviews related to dining experiences within the same urban area~\cite{rayana2015collective}. Expanding the geographical coverage, YelpNYC encompasses restaurant reviews from New York City, preserving the intrinsic data architecture but diversifying in terms of geographic and urban dynamics~\cite{rayana2015collective}. Diverging from the service industry focus, the Amazon dataset contains user reviews from the e-commerce sector~\cite{kaghazgaran2018combating}. Users are identified as fraudulent if they have reviewed multiple products involved in crowd-sourcing campaigns. 
\begin{table}[t]
    \caption{Summary statistics for datasets used in experiments.} 

    \centering
    \resizebox{\linewidth}{!}{
    \begin{tabular}{c | c  | c  | c  | c  } 
    \hline
    
    Datasets  & \# nodes & \# edges  & \# attributes & \# anomalies  \\ \hline

   YelpHotel & 4322 & 101,800 & 8000 & 250 \\
   YelpRes & 5012 & 355,144 & 8000 & 250 \\
   YelpNYC & 21,040 & 1,658,137 & 10000 & 1000 \\
   Amazon & 17,496 & 495,213 & 10000 & 705 \\ \hline

    \end{tabular}}
    \label{table: data statistics}
\end{table}

\begin{table*}[t]
    \caption{one-shot performance comparison with the existing GAD methods in terms of AUC-ROC and AUC-PR. The best results are marked in blue.}
    \centering
    \resizebox{\linewidth}{!}{
    \begin{tabular}{c | c c | c c | c c | c c } 
    \hline
    
    Datasets & \multicolumn{2}{c|}{YelpHotel$\rightarrow$YelpNYC} & \multicolumn{2}{c|}{YelpNYC$\rightarrow$YelpHotel} & \multicolumn{2}{c|}{YelpHotel$\rightarrow$YelpRes} & \multicolumn{2}{c}{YelpRes$\rightarrow$YelpHotel} \\  \hline

    Evaluation Metrics & AUC-ROC & AUC-PR & AUC-ROC & AUC-PR & AUC-ROC & AUC-PR & AUC-ROC & AUC-PR \\ \hline

    SemiGNN (ICDM 2019) ~\cite{wang2019semi}  &  0.624$\pm$0.012 &  0.133$\pm$0.025 &  0.737$\pm$0.025 &  0.200$\pm$0.033 &  0.684$\pm$0.056 &  0.131$\pm$0.088 &  0.737$\pm$0.025 &  0.200$\pm$0.033 \\
    
    ANEMONE (CIKM 2021)~\cite{jin2021anemone} &  0.437$\pm$0.015 &  0.039$\pm$0.002 &  0.457$\pm$0.023 &  0.051$\pm$0.003 &  0.463$\pm$0.069 &  0.043$\pm$0.005 &  0.457$\pm$0.023 &  0.051$\pm$0.003 \\
    
    meta-GDN (WWW 2021)~\cite{ding2021few}  &  0.840$\pm$0.057 &  0.360$\pm$0.065 &  0.688$\pm$0.055 &  0.219$\pm$0.032 &  0.949$\pm$0.023 &  0.564$\pm$0.148 &  0.687$\pm$0.057 &  0.235$\pm$0.042 \\
    
    ANEMONE-FS (arXiv 2022) ~\cite{zheng2022unsupervised} &  0.474$\pm$0.001 &  0.078$\pm$0.000 &  0.752$\pm$0.002 &  0.239$\pm$0.002 &  0.446$\pm$0.007 &  0.042$\pm$0.001 &  0.752$\pm$0.002 &  0.239$\pm$0.002 \\
    
    COLA (TNNLS 2022)~\cite{liu2021anomaly} &  0.670$\pm$0.009 &  0.083$\pm$0.004 &  0.495$\pm$0.027 &  0.064$\pm$0.006 &  0.626$\pm$0.025 &  0.064$\pm$0.005 &  0.495$\pm$0.027 &  0.064$\pm$0.006 \\
    
    ACT (AAAI 2023)~\cite{wang2023cross}  &  0.833$\pm$0.001 &  0.269$\pm$0.009 &  0.790$\pm$0.006 &  0.224$\pm$0.007 &  0.901$\pm$0.012 &  0.301$\pm$0.021 &  0.785$\pm$0.017	&0.218$\pm$0.009 \\
    
    PREM (ICDM 2023)~\cite{pan2023prem}   &  0.557$\pm$0.012 &  0.054$\pm$0.002 &  0.662$\pm$0.025 &  0.111$\pm$0.014 &  0.498$\pm$0.055 &  0.048$\pm$0.007 &  0.662$\pm$0.025 &  0.111$\pm$0.014 \\
    
    SL-GAD (TKDE 2023)~\cite{zheng2021generative} &  0.699$\pm$0.002 &  0.152$\pm$0.001 &  0.601$\pm$0.013 &  0.173$\pm$0.008 &  0.753$\pm$0.002 &  0.204$\pm$0.002 &  0.601$\pm$0.013 &  0.173$\pm$0.008 \\
    
    CDFS-GAD  (ours)  &  \textcolor{blue}{\textbf{0.864$\pm$0.008}} &  \textcolor{blue}{\textbf{0.361$\pm$0.028}}  
        & \textcolor{blue}{\textbf{0.796$\pm$0.017}} &  \textcolor{blue}{\textbf{0.264$\pm$0.041}} 
    &  \textcolor{blue}{\textbf{0.992$\pm$0.005}} &  \textcolor{blue}{\textbf{0.903$\pm$0.054}} &  \textcolor{blue}{\textbf{0.798$\pm$0.021}} &  \textcolor{blue}{\textbf{0.268$\pm$0.016}} \\ \hline

    & \multicolumn{2}{c|}{YelpHotel$\rightarrow$Amazon} & \multicolumn{2}{c|}{Amazon$\rightarrow$YelpHotel} & \multicolumn{2}{c|}{YelpNYC$\rightarrow$YelpRes} & \multicolumn{2}{c}{YelpRes$\rightarrow$YelpNYC} \\ \hline

    & AUC-ROC & AUC-PR & AUC-ROC & AUC-PR & AUC-ROC & AUC-PR & AUC-ROC & AUC-PR \\ \hline

    SemiGNN (ICDM 2019) ~\cite{wang2019semi}  &  0.374$\pm$0.091 &  0.040$\pm$0.013 &  0.737$\pm$0.025 &  0.200$\pm$0.033 &  0.684$\pm$0.056 &  0.131$\pm$0.088 &  0.624$\pm$0.012 &  0.133$\pm$0.025 \\
    
    ANEMONE (CIKM 2021)~\cite{jin2021anemone}  &  0.585$\pm$0.026 &  0.044$\pm$0.003 &  0.457$\pm$0.023 &  0.051$\pm$0.003 &  0.463$\pm$0.069 &  0.043$\pm$0.005 &  0.437$\pm$0.015 &  0.039$\pm$0.002 \\
    
    meta-GDN (WWW 2021)~\cite{ding2021few}  &  0.772$\pm$0.098 &  0.221$\pm$0.085 &  0.677$\pm$0.029 &  0.244$\pm$0.044 &  0.950$\pm$0.029 &  0.603$\pm$0.147 &  0.853$\pm$0.076 &  0.386$\pm$0.098 \\
    
    ANEMONE-FS (arXiv 2022) ~\cite{zheng2022unsupervised} &  0.486$\pm$0.002 &  0.036$\pm$0.000 &  0.752$\pm$0.002 &  0.239$\pm$0.002 &  0.446$\pm$0.007 &  0.042$\pm$0.001 &  0.474$\pm$0.001 &  0.078$\pm$0.000 \\
    
    COLA (TNNLS 2022)~\cite{liu2021anomaly}  &  0.593$\pm$0.004 &  0.045$\pm$0.000 &  0.495$\pm$0.027 &  0.064$\pm$0.006 &  0.626$\pm$0.025 &  0.064$\pm$0.005 &  0.670$\pm$0.009 &  0.083$\pm$0.004 \\
    
    ACT (AAAI 2023)~\cite{wang2023cross}  &  0.638$\pm$0.003 &  0.052$\pm$0.001 
    &  0.788$\pm$0.028 &  0.219$\pm$0.007
    &  0.932$\pm$0.003 &  0.361$\pm$0.011 &  0.832$\pm$0.006 &  0.270$\pm$0.012 \\
    
    PREM (ICDM 2023)~\cite{pan2023prem}   &  0.511$\pm$0.007 &  0.040$\pm$0.001 &  0.662$\pm$0.025 &  0.111$\pm$0.014 &  0.498$\pm$0.055 &  0.048$\pm$0.007 &  0.557$\pm$0.012 &  0.054$\pm$0.002 \\
    
    SL-GAD (TKDE 2023)~\cite{zheng2021generative}   &  0.832$\pm$0.008 &  \textcolor{blue}{\textbf{0.319$\pm$0.014}} &  0.601$\pm$0.013 &  0.173$\pm$0.008 &  0.753$\pm$0.002 &  0.204$\pm$0.002 &  0.699$\pm$0.002 &  0.152$\pm$0.001 \\
    
    CDFS-GAD  (ours)  &  \textcolor{blue}{\textbf{0.835$\pm$0.062}} &  0.271$\pm$0.094 
    &  \textcolor{blue}{\textbf{0.789$\pm$0.014}} &  \textcolor{blue}{\textbf{0.246$\pm$0.027}} 
    &  \textcolor{blue}{\textbf{0.991$\pm$0.004}} &  \textcolor{blue}{\textbf{0.885$\pm$0.057}} 
    & \textcolor{blue}{\textbf{0.872$\pm$0.005}} &  \textcolor{blue}{\textbf{0.387$\pm$0.017}}
    \\ \hline

     & \multicolumn{2}{c|}{YelpNYC$\rightarrow$Amazon} & \multicolumn{2}{c|}{Amazon$\rightarrow$YelpNYC} & \multicolumn{2}{c|}{YelpRes$\rightarrow$Amazon} & \multicolumn{2}{c}{Amazon$\rightarrow$YelpRes} \\ \hline
    
    & AUC-ROC & AUC-PR & AUC-ROC & AUC-PR & AUC-ROC & AUC-PR & AUC-ROC & AUC-PR \\  \hline

    SemiGNN (ICDM 2019) ~\cite{wang2019semi}  &  0.374$\pm$0.091 &  0.040$\pm$0.013 &  0.624$\pm$0.012 &  0.133$\pm$0.025 &  0.374$\pm$0.091 &  0.040$\pm$0.013 &  0.684$\pm$0.056 &  0.131$\pm$0.088 \\
    
    ANEMONE (CIKM 2021)~\cite{jin2021anemone}  &  0.585$\pm$0.026 &  0.044$\pm$0.003 &  0.437$\pm$0.015 &  0.039$\pm$0.002 &  0.585$\pm$0.026 &  0.044$\pm$0.003 &  0.463$\pm$0.069 &  0.043$\pm$0.005 \\
    
    meta-GDN (WWW 2021)~\cite{ding2021few}   &  \textcolor{blue}{\textbf{0.863$\pm$0.058}} &  \textcolor{blue}{\textbf{0.351$\pm$0.062}} &  0.865$\pm$0.026 &  \textcolor{blue}{\textbf{0.397$\pm$0.108}} &  0.753$\pm$0.131 &  0.224$\pm$0.121 &  0.953$\pm$0.021 &  0.604$\pm$0.133 \\
    
    ANEMONE-FS (arXiv 2022)~\cite{zheng2022unsupervised} &  0.486$\pm$0.002 &  0.036$\pm$0.000 &  0.474$\pm$0.001 &  0.078$\pm$0.000 &  0.486$\pm$0.002 &  0.036$\pm$0.000 &  0.446$\pm$0.007 &  0.042$\pm$0.001 \\
    
    COLA (TNNLS 2022)~\cite{liu2021anomaly}  &  0.593$\pm$0.004 &  0.045$\pm$0.000 &  0.670$\pm$0.009 &  0.083$\pm$0.004 &  0.593$\pm$0.004 &  0.045$\pm$0.000 &  0.626$\pm$0.025 &  0.064$\pm$0.005 \\
    
    ACT (AAAI 2023)~\cite{wang2023cross}  &  0.641$\pm$0.007 &  0.052$\pm$0.001 &  0.834$\pm$0.002 &  0.278$\pm$0.007 &  0.629$\pm$0.004 &  0.050$\pm$0.001 &  0.912$\pm$0.011 &  0.315$\pm$0.023 \\
    
    PREM (ICDM 2023)~\cite{pan2023prem} &  0.511$\pm$0.007 &  0.040$\pm$0.001 &  0.557$\pm$0.012 &  0.054$\pm$0.002 &  0.511$\pm$0.007 &  0.040$\pm$0.001 &  0.498$\pm$0.055 &  0.048$\pm$0.007 \\
    
    SL-GAD (TKDE 2023)~\cite{zheng2021generative}  &  0.832$\pm$0.008 &  0.319$\pm$0.014 &  0.699$\pm$0.002 &  0.152$\pm$0.001 &  0.832$\pm$0.008 &  \textcolor{blue}{\textbf{0.319$\pm$0.014}} &  0.753$\pm$0.002 &  0.204$\pm$0.002 \\
    
    CDFS-GAD  (ours)  &  0.784$\pm$0.075 &  0.200$\pm$0.107 &  \textcolor{blue}{\textbf{0.865$\pm$0.007}} &  0.366$\pm$0.032 & \textcolor{blue}{\textbf{0.855$\pm$0.030}} &  0.308$\pm$0.061 & \textcolor{blue}{\textbf{ 0.991$\pm$0.006}} &  \textcolor{blue}{\textbf{0.903$\pm$0.059}} \\ \hline
    
    \end{tabular}}
    \label{tab: comparison study}
\end{table*}

Following similar settings in~\cite{ding2021cross}~\cite{wang2023cross}, we have constructed twelve cross-domain pairs by pairing each dataset with every other. These pairs display varying degrees of domain similarities and dissimilarities. For instance, while YelpNYC and YelpRes both contain user reviews related to restaurants, they differ in their geographic location. YelpHotel and YelpRes, while both reside in Chicago, cater to distinct service sectors—accommodation and dining. The shift from YelpHotel to YelpNYC is relatively larger as they span not just the service type but also the urban environment. Nevertheless, their anomaly detection methodologies are similar as unified by Yelp's filtering algorithm. The progression from YelpNYC to Amazon presents a relatively larger domain transition, moving from restaurant reviews to e-commerce. This transition engages diverse user interaction dynamics and review behaviors.

\subsection{Implementation Details}

The proposed CDFS-GAD framework is implemented via PyTorch, and Adam optimizer~\cite{kingma2014adam} with a learning rate of 0.0005 is used for model training. The architecture of the backbone encoder is composed of a two-layer GraphSAGE with dimensionalities of 256 and 64, respectively. Within the contrastive learning module, the edge-dropping probability $p$ is set as 0.1, and the balance weight $\alpha$ is set to 0.5. The readout function $\mathcal{R}(\cdot)$ is chosen to be the simple average over the input node representations. For the self-training phase, the parameters $\beta_1$ and $\beta_2$ are chosen to be 0.02 and 0.25, respectively. In the prompt learning module, we set the number of bases $m$ to 5 for both source and target graphs. The training epochs for the joint training stage and the self-training stage are set to 50 and 100 epochs, respectively.

For the empirical evaluation, we partition the nodes of the target graph into three segments: $40\%$ for training, $20\%$ for validation, and the remaining $40\%$ for testing purposes. Additionally, the reported results are averaged over five independent trials with each initialized with different random seeds.

\subsection{Baselines and Evaluation Metrics}

We compare eight recent state-of-the-art methods spanning several categories: unsupervised GAD methods (COLA, SL-GAD, ANEMONE, and PREM), semi-supervised GAD (SemiGNN), few-shot GAD methods (meta-GDN and ANEMONE-FS) and cross-domain GAD (ACT). The unsupervised GAD methods leverage solely the unlabeled data from the target domain, while cross-domain GAD utilizes both the labeled data from the source domain and the unlabeled data from the target domain. SemiGNN operates exclusively on the labeled data of the target domain. Similarly, AENMONE-FS focuses on the labeled data in the target domain. For meta-GDN, we follow a similar setting in its original implementation by splitting the source domain graph into disjoint graphs, and then we treat those sub-graphs as the auxiliary networks and the target domain graph as the target network.

We adopt two widely used metrics for performance evaluation: the area under the receiver operating characteristic curve (AUC-ROC) and the area under the precision-recall curve (AUC-PR). These metrics quantify the anomaly detection performance over a diverse range of decision thresholds.
\begin{figure*}[t]
\begin{subfigure}[b]{1\linewidth}
    \centering
    \includegraphics[height=0.8cm]{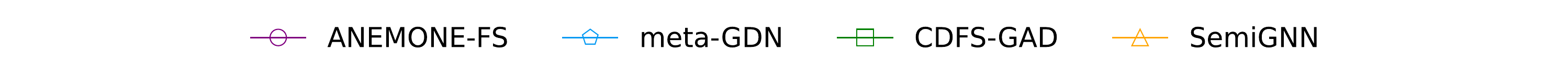}
\end{subfigure}
\newline
\begin{subfigure}[b]{0.33\linewidth}
    \centering
    \includegraphics[height=3.3cm]{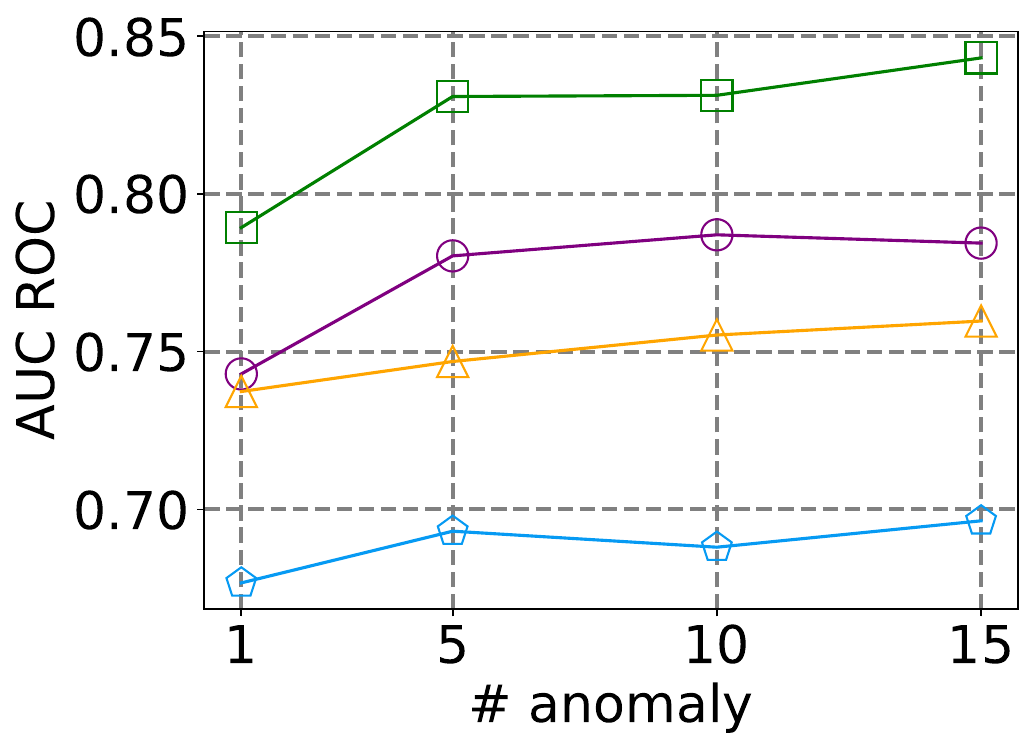}
    \caption{Amazon$\rightarrow$YelpHotel}
     \label{fig: amz-to-htl fewshot}
\end{subfigure}
\begin{subfigure}[b]{0.33\linewidth}
    \centering
    \includegraphics[height=3.3cm]{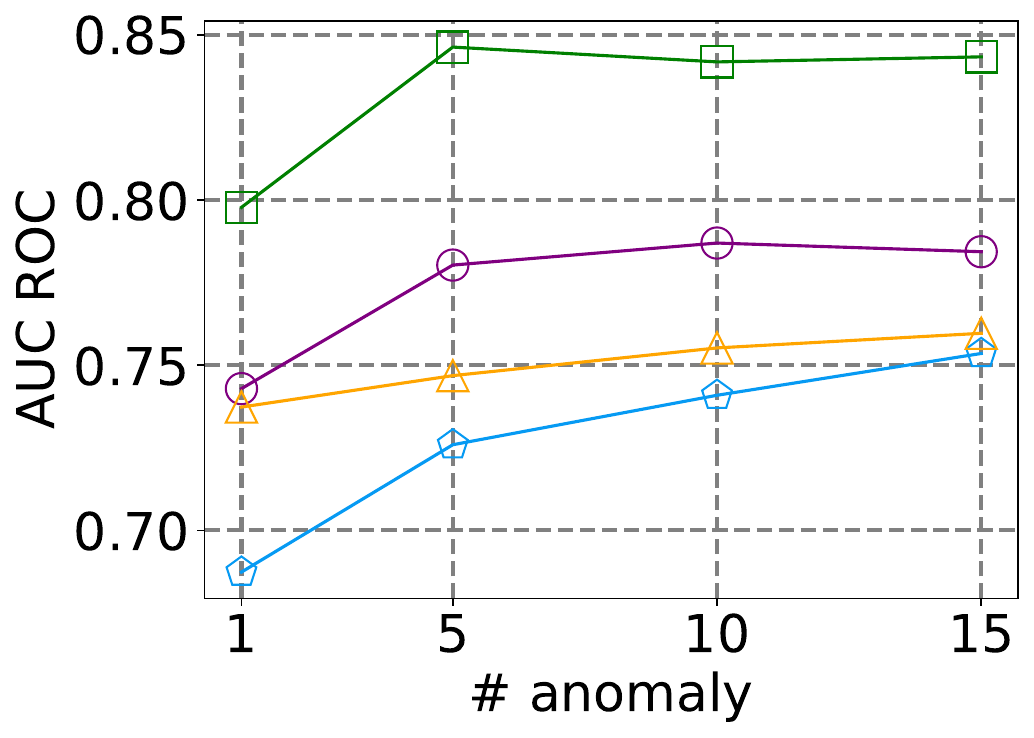} 
    \caption{YelpRes$\rightarrow$YelpHotel}
    \label{fig: res-to-amz fewshot}
\end{subfigure}
\begin{subfigure}[b]{0.33\linewidth}
    \centering
    \includegraphics[height=3.3cm]{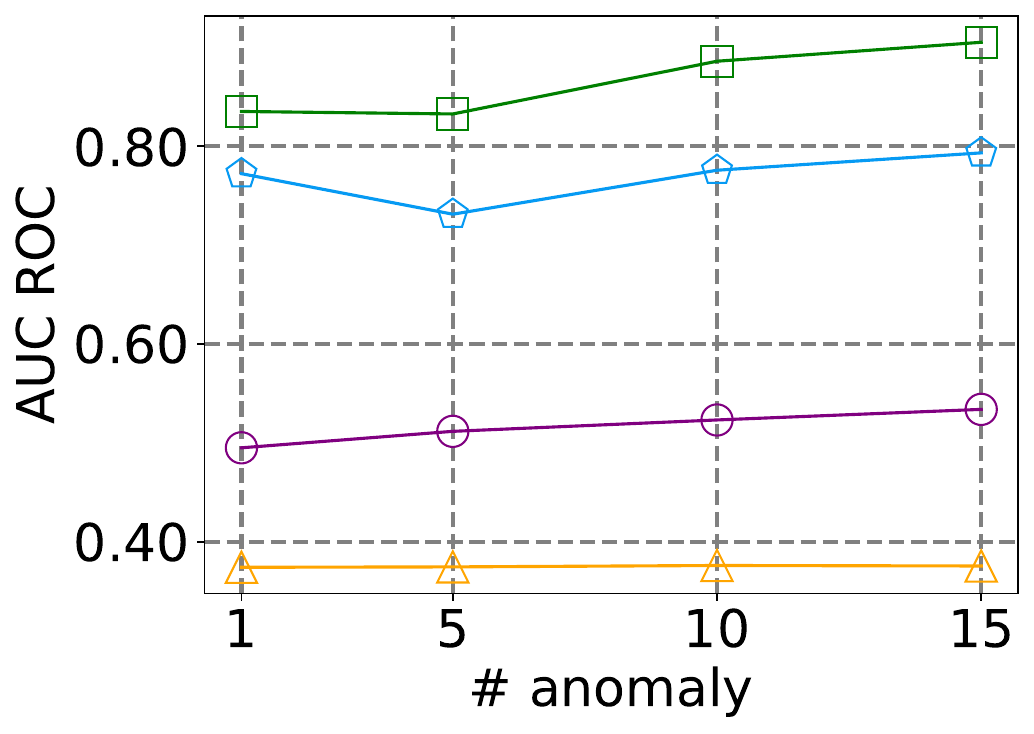} 
    \caption{YelpHotel$\rightarrow$Amazon}
    \label{fig: res-to-amz fewshot}
\end{subfigure}
\newline
\begin{subfigure}[b]{0.33\linewidth}
    \centering
    \includegraphics[height=3.3cm]{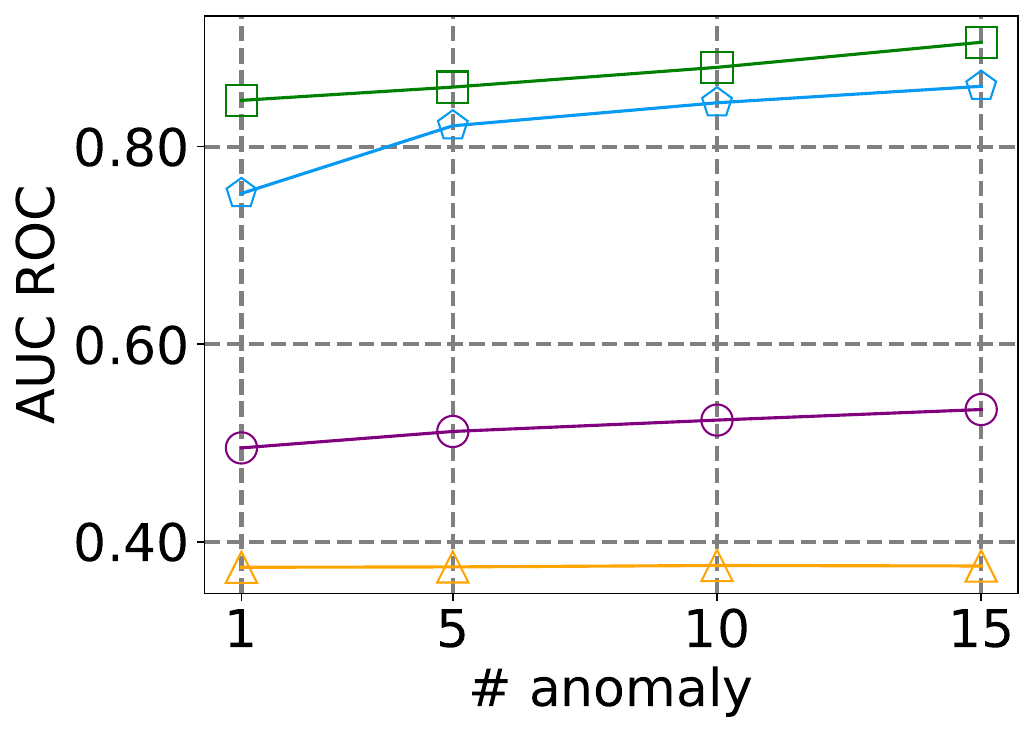} 
    \caption{YelpRes$\rightarrow$Amazon}
    \label{fig: res-to-amz fewshot}
\end{subfigure}
\begin{subfigure}[b]{0.33\linewidth}
    \centering
    \includegraphics[height=3.3cm]{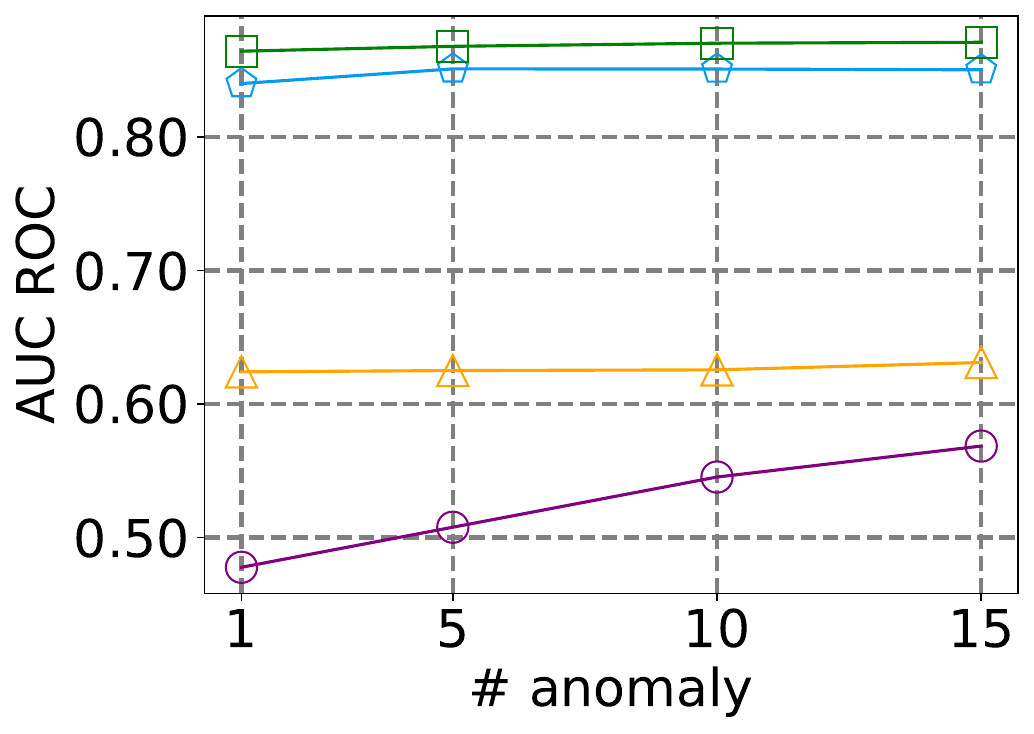} 
    \caption{YelpHotel$\rightarrow$YelpNYC}
    \label{fig: htl-to-nyc fewshot}
\end{subfigure}
\begin{subfigure}[b]{0.33\linewidth}
    \centering
    \includegraphics[height=3.3cm]{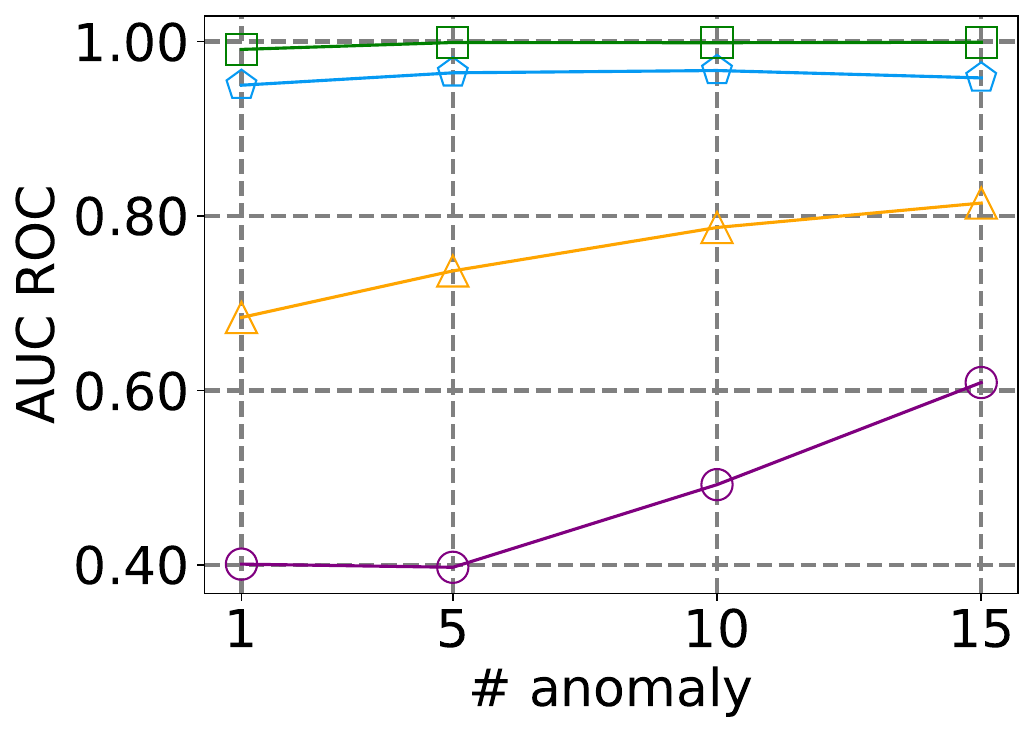} 
    \caption{YelpNYC$\rightarrow$YelpRes}
    \label{fig: nyc-to-res fewshot}
\end{subfigure}
\caption{Various $K$-shot performance comparison with the state-of-the-art GAD methods in terms of AUC-ROC.}
\label{fig: few-shot}
\end{figure*}

\subsection{Comparison Study}

Table~\ref{tab: comparison study} presents the one-shot performance of all models across 12 data pairs in terms of AUC-ROC and AUC-PR. Predominantly, our model outperforms other state-of-the-art methods across most data pairs. In particular, the improvements are evident in pairs involving YelpNYC, YelpRes, and YelpHotel, which share relatively larger domain similarities. For example, YelpRes and YelpHotel contain data collected under the same urban context, while YelpNYC and YelpRes belong to the same service industry. Conversely, our model experiences a marginal performance decline with pairs that include the Amazon dataset, likely due to the discrepancy in the anomaly generation mechanism and service industry between the Amazon and Yelp datasets. 

In comparison to meta-GDN, a competing method that also utilizes the label from two domains, our model surpasses it by a noticeable margin. We attribute this enhanced performance to our inclusion of domain-adaptive components that are adept at capturing domain-invariant and domain-specific information. In contrast, the performance of meta-GDN suffers due to its assumption that auxiliary and target networks belong to the same domain—an assumption that does not hold in our cross-domain setup and leads to its underperformance.

Regarding other methods, the unsupervised GAD methods generally underperform compared to the cross-domain unsupervised GAD method. ACT benefits from incorporating labeled information from the source domain, which enhances its learning capabilities for the target domain. On the other hand, the few-shot GAD method meta-GDN on average performs better than the unsupervised GAD methods and ACT, which is likely due to its access to additional labeled data from the target domain. However, methods like SemiGNN and ANEMONE-FS, which also rely on few-shot labels in the target domain, do not consistently perform well. Their underperformance relative to meta-GDN and even some unsupervised methods can be primarily attributed to the constraints of their loss functions (e.g., cross-entropy), which are less robust in extremely label-scarce environments than the few-shot enhanced deviation network employed by meta-GDN.

In summary, the superiority of our model to other methods can be attributed to the integration of the domain-adaptive modules that accommodate the similarity and difference between two domains, augmented by specialized anomaly detection loss and self-training for enhanced learning under few-shot scenarios. These elements collectively contribute to the final performance by effectively leveraging cross-domain information and sparsely labeled data.
\begin{table*}[t]
    \caption{Ablation study under the one-shot task in terms of AUC-ROC and AUC-PR. The best results are marked in blue. $w/o$ denotes without a certain module.}
    \centering
    \resizebox{\linewidth}{!}{
    \begin{tabular}{c | c c | c c | c c | c c } 
    \hline
    
    Datasets & \multicolumn{2}{c|}{YelpRes$\rightarrow$YelpHotel} & \multicolumn{2}{c|}{YelpNYC$\rightarrow$YelpHotel} & \multicolumn{2}{c|}{YelpRes$\rightarrow$Amazon} & \multicolumn{2}{c}{YelpNYC$\rightarrow$YelpRes} \\  \hline

    Evaluation Metrics & AUC-ROC & AUC-PR & AUC-ROC & AUC-PR & AUC-ROC & AUC-PR & AUC-ROC & AUC-PR \\ \hline

    $w/o$ domain-specific prompt  &  0.749$\pm$0.030 &  0.228$\pm$0.033 &  0.760$\pm$0.028 &  0.239$\pm$0.042 &  0.822$\pm$0.021 &  0.250$\pm$0.029 &  0.988$\pm$0.003 &  0.882$\pm$0.060 \\

    $w/o$ $\LL_{intra}$ &  0.789$\pm$0.025 &  0.265$\pm$0.041 &  0.784$\pm$0.021 &  0.257$\pm$0.031 &  0.819$\pm$0.076 &  0.265$\pm$0.104 &  0.984$\pm$0.004 &  0.838$\pm$0.106 \\

    $w/o$ $\LL_{intra}$ + $\LL_{inter}$  &  0.780$\pm$0.022 &  0.232$\pm$0.038 &  0.782$\pm$0.012 &  0.228$\pm$0.034 &  0.814$\pm$0.086 &  0.259$\pm$0.125 &  0.980$\pm$0.004 &  0.783$\pm$0.086 \\

    $\LL_{hsc}$ only  &  0.787$\pm$0.024 &  0.249$\pm$0.032 &  0.791$\pm$0.016 &  0.246$\pm$0.024 &  0.840$\pm$0.029 &  0.266$\pm$0.063 &  0.984$\pm$0.008 &  0.809$\pm$0.082 \\

    $w/o$ source domain supervision 
    & 0.769$\pm$0.013  & 0.219$\pm$0.025 
    & 0.763$\pm$0.015 & 0.205$\pm$0.024
    & 0.798$\pm$0.067 & 0.221$\pm$0.078 & 0.982$\pm$0.003 & 0.800$\pm$0.039 \\ 

    $w/o$ self-training  &  0.788$\pm$0.013 &  0.240$\pm$0.025 &  0.791$\pm$0.011 &  0.241$\pm$0.021 &  0.796$\pm$0.047 &  0.257$\pm$0.102 &  0.962$\pm$0.012 &  0.649$\pm$0.099 \\
    
    CDFS-GAD  &  \textcolor{blue}{\textbf{0.798$\pm$0.021}} &  \textcolor{blue}{\textbf{0.268$\pm$0.016}} &  \textcolor{blue}{\textbf{0.796$\pm$0.017}} &  \textcolor{blue}{\textbf{0.264$\pm$0.041}} &  \textcolor{blue}{\textbf{0.855$\pm$0.030}} &  \textcolor{blue}{\textbf{0.308$\pm$0.061}} &  \textcolor{blue}{\textbf{0.991$\pm$0.004}} &  \textcolor{blue}{\textbf{0.885$\pm$0.057}} \\ \hline

    \end{tabular}}
    \label{table: ablation_study}
\end{table*}

\begin{figure*}[t!]
\begin{subfigure}[b]{1\linewidth}
    \centering
    \includegraphics[height=0.8cm]{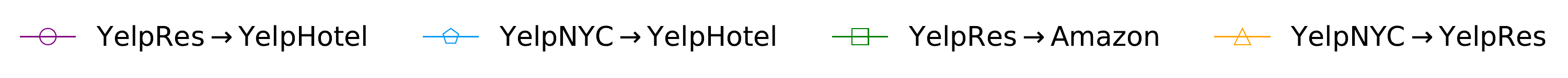}
\end{subfigure}
\newline
\begin{subfigure}[b]{0.24\linewidth}
    \centering
    \includegraphics[height=3.5cm]{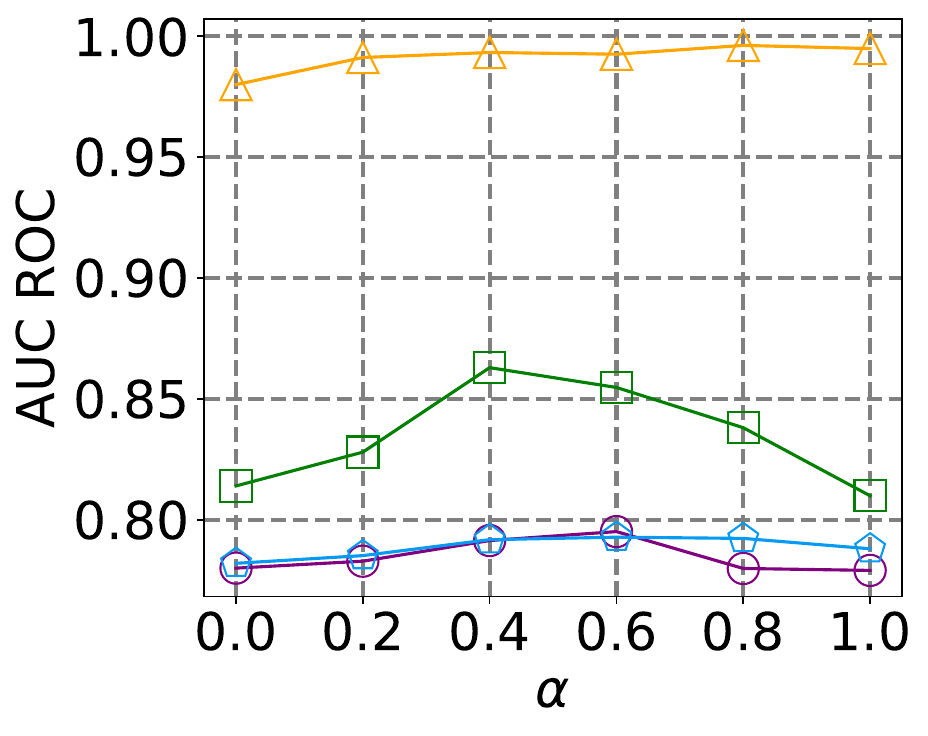} 
    \caption{Balance Weight $\alpha$}
    \label{fig: balance_weight}
\end{subfigure}
\begin{subfigure}[b]{0.24\linewidth}
    \centering
    \includegraphics[height=3.5cm]{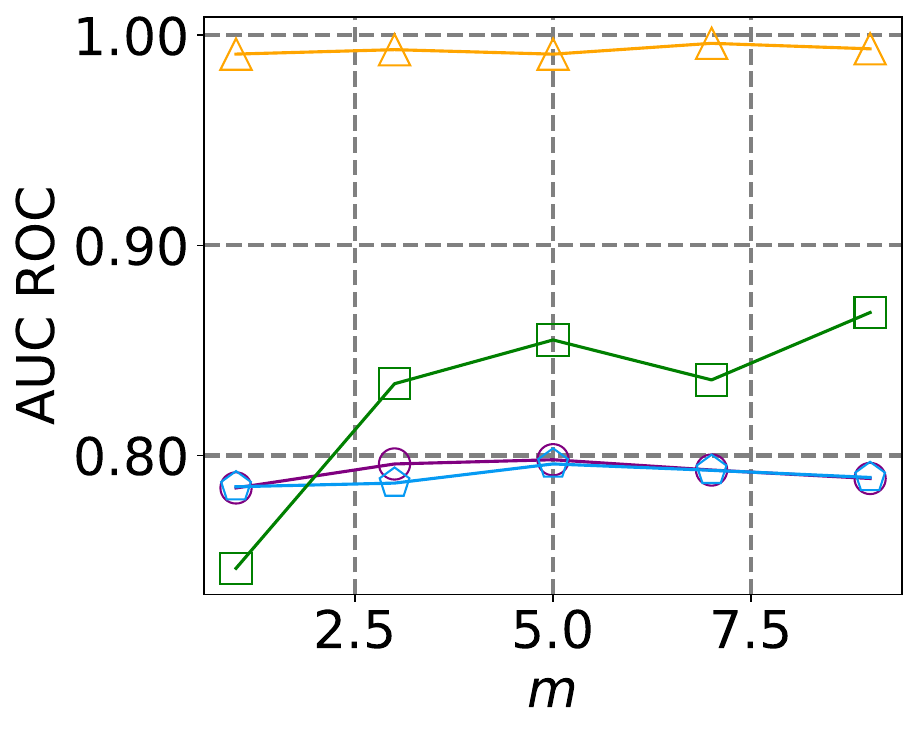} 
    \caption{Prompt Basis Number $m$}
    \label{fig: anomaly_ratio}
\end{subfigure}
\begin{subfigure}[b]{0.24\linewidth}
    \centering
    \includegraphics[height=3.5cm]{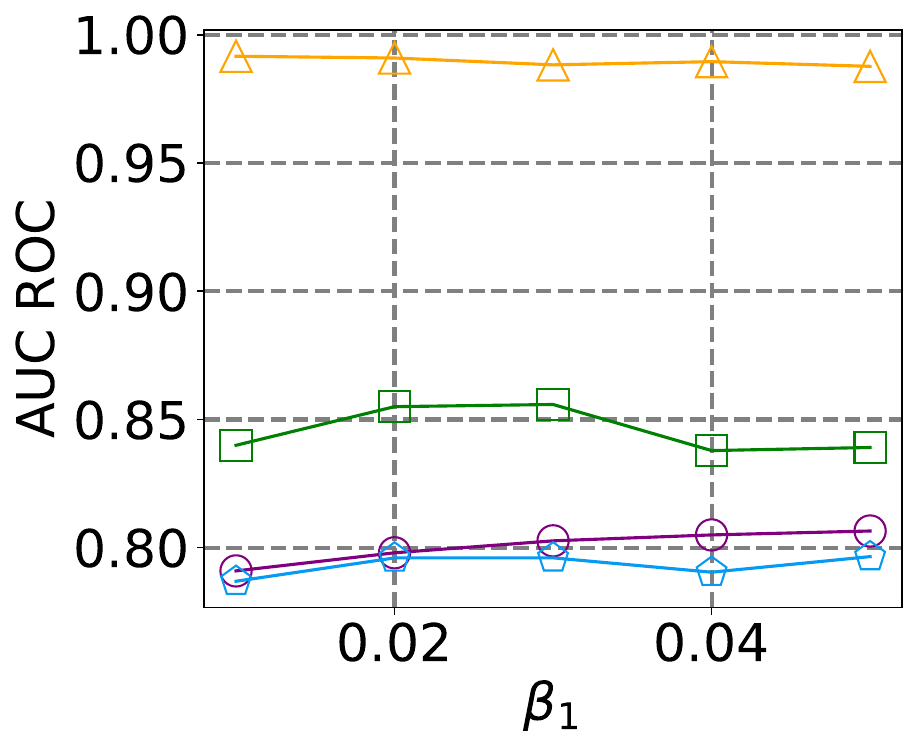} 
    \caption{Anomaly Ratio $\beta_1$}
    \label{fig: anomaly_ratio}
\end{subfigure}
\begin{subfigure}[b]{0.24\linewidth}
    \centering
    \includegraphics[height=3.5cm]{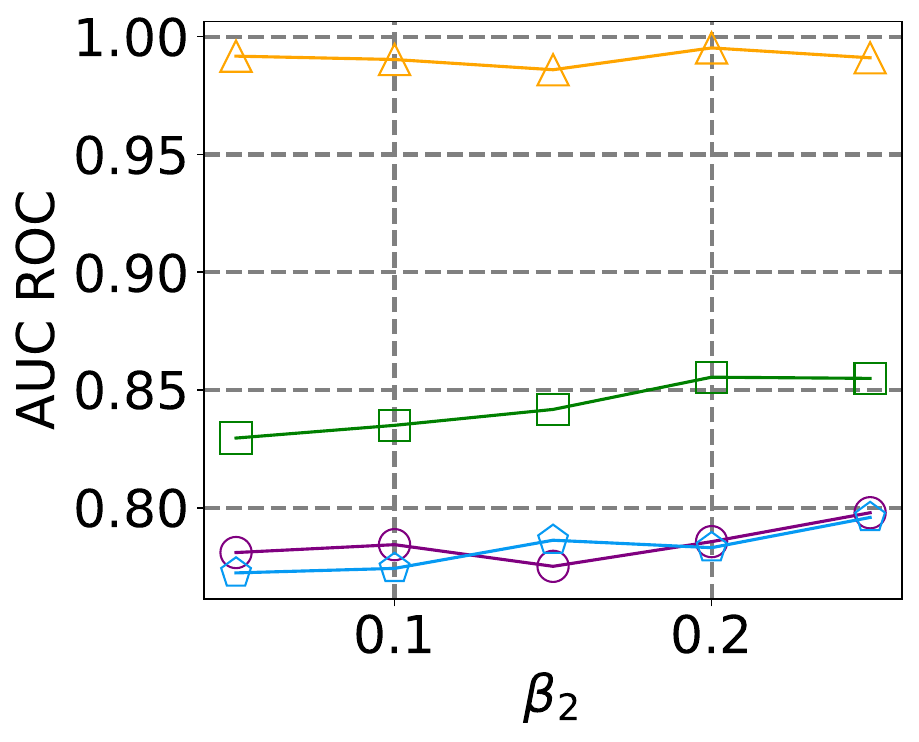} 
    \caption{Normal Ratio $\beta_2$}
    \label{fig: normal_ratio}
\end{subfigure}
\caption{Sensitivity analysis of the hyper-parameters $\alpha$, $m$, $\beta_1$, and $\beta_2$ of CDFS-GAD under the one-shot task in terms of AUC-ROC.}
\label{fig: sensitivity}
\end{figure*}

\subsection{Evaluation under Different Shots}

Figure~\ref{fig: few-shot} presents the $K$-shot performance of semi-supervised and few-shot GAD methods in terms of AUC-ROC. We can observe that there is a large discrepancy in performance between SemiGNN and ANEMONE-FS, which rely solely on the target domain, and those like CDFS-GAD and meta-GDN which utilize the label information from both domains. The performance of SemiGNN and ANEMONE-FS lags significantly behind, particularly in scenarios with extremely limited labeled data (e.g., one-shot). In contrast, CDFS-GAD and meta-GDN show relative stability across different shots. This stability can be attributed to two factors: firstly, the inclusion of cross-domain labeled data provides a richer set of examples for model training. Secondly, the anomaly detection frameworks in CDFS-GAD and meta-GDN are tailored for few-shot scenarios, i.e., hypersphere classification or deviation network, thereby improving their performance when labeled data is scarce. Despite the strengths of meta-GDN, our model consistently outperforms it across all $K$-shot settings. Specifically, CDFS-GAD is able to extract the relevant features from the source domain while retaining the unique characteristics of the target graph anomaly distribution, thereby improving the model generalization in the target domain.

\subsection{Ablation Study}

We conduct an ablation study for one-shot settings according to the following variants:

\begin{itemize}

    \item $w/o$ domain-specific prompt: Removing the domain-specific prompt module, which is added to the hidden representations generated from the GNN encoder.
    
    \item $w/o$ $\LL_{intra}$: Removing the intra-domain contrastive loss.
    
    \item $w/o$ $\LL_{intra} + \LL_{inter}$: Removing both the intra-domain and inter-domain contrastive loss.
    
    \item $\LL_{hsc}$ only: Replacing the $\LL_{dahsc}$ with $\LL_{hsc}$, meaning that we adopted separate anomaly detectors that rely on different learnable centers for the source and target domain, respectively.
    
    \item $w/o$ source domain supervision: We train the model based on purely the labeled data from the target domain. 
    
    \item $w/o$ self-training: We use the prediction results generated purely from joint learning based on the labeled data from both the source and target domain.
    
\end{itemize}

We summarize our findings according to the results presented in Table~\ref{table: ablation_study}:
First, the omission of the domain-specific prompt learning module results in a notable performance deduction. This suggests that while capturing domain-invariant information is critical, the domain-specific attributes are also important, as part of the normal/abnormal node behaviors can still be tied to their respective domains uniquely. Secondly, the drop in performance upon removal of the intra-contrastive learning component suggests that aligning source and target feature representations is vital for transferring knowledge between two domains. The absence of this module would impair the ability of our model to extract domain-invariant attributes between the two domains. In contrast, inter-contrastive learning helps to improve the representation quality of normal nodes within each domain, removing it along with intra-contrastive learning would impede the ability of our model to effectively represent features that are common across and unique to each domain. Moreover, replacing $\LL_{dahsc}$ with distinct $\LL_{hsc}$ for each domain also leads to a decline in performance. The inclusion of a domain-invariant center in addition to the domain-specific center improves the flexibility of our model to identify anomalies based on both within-domain and cross-domain information. Additionally, there is a notable drop in performance when removing source domain supervision. This agrees with our assumptions that labeled data from the source domain contributes to learning a more accurate model for the target domain. Lastly, the performance downturn in the absence of self-training indicates its critical role in addressing the high uncertainty associated with few-shot scenarios. This process helps mitigate overfitting by retraining the model on its most confident predictions.

\begin{figure*}[ht]
\begin{subfigure}[b]{0.33\linewidth}
    \centering
    \includegraphics[height=4.5cm]{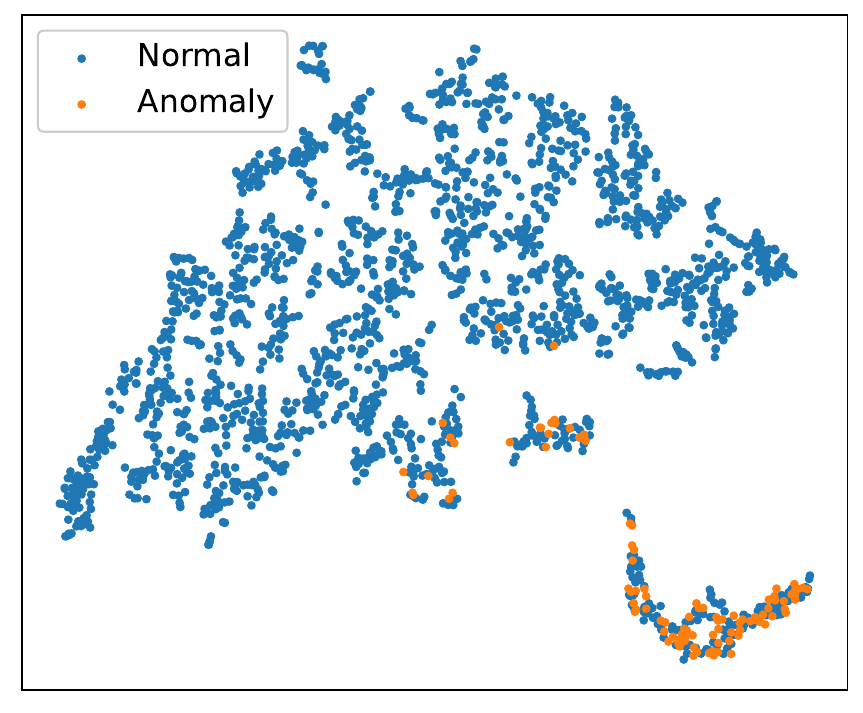} 
    \caption{YelpHTL$\rightarrow$YelpRes (meta-GDN)}
    \label{fig: res-to-amz fewshot}
\end{subfigure}
\begin{subfigure}[b]{0.33\linewidth}
    \centering
    \includegraphics[height=4.5cm]{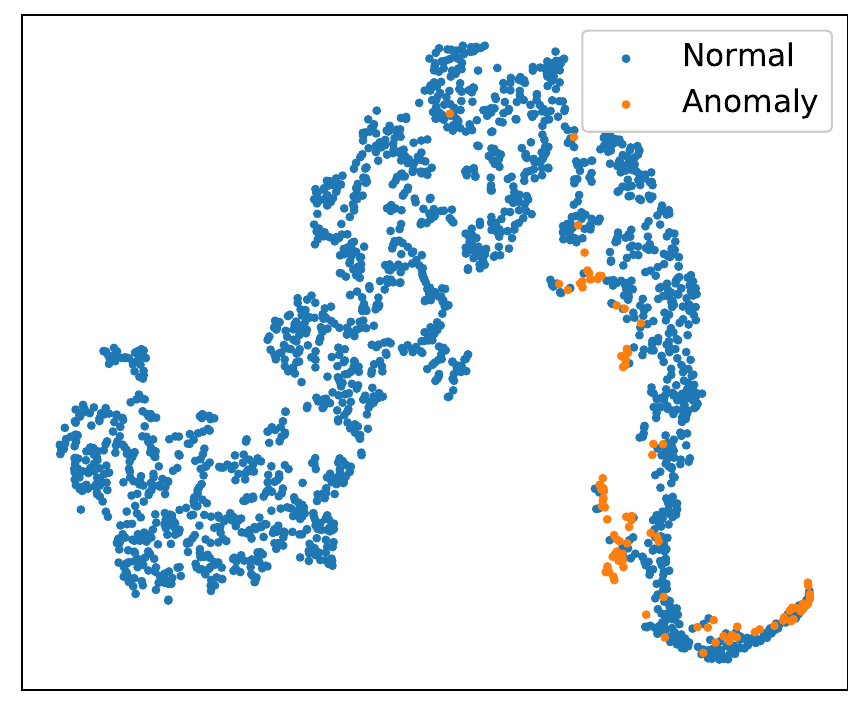} 
    \caption{YelpHTL$\rightarrow$YelpRes (ACT)}
    \label{fig: res-to-amz fewshot}
\end{subfigure}
\begin{subfigure}[b]{0.33\linewidth}
    \centering
    \includegraphics[height=4.5cm]{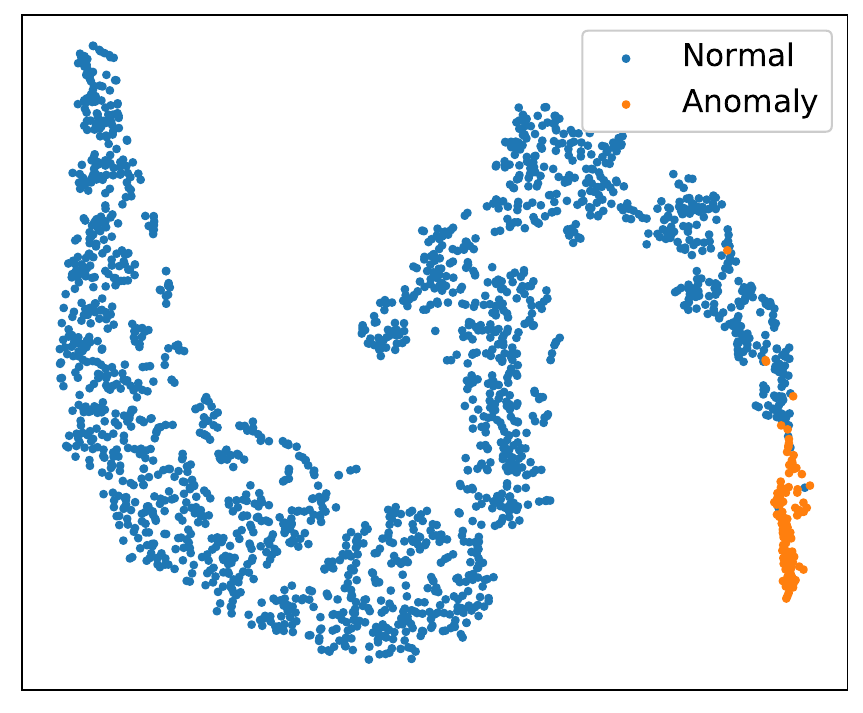} 
    \caption{YelpHTL$\rightarrow$YelpRes (CDFS-GAD)}
    \label{fig: nyc-to-res fewshot}
\end{subfigure}
\newline
\begin{subfigure}[b]{0.33\linewidth}
    \centering
    \includegraphics[height=4.5cm]{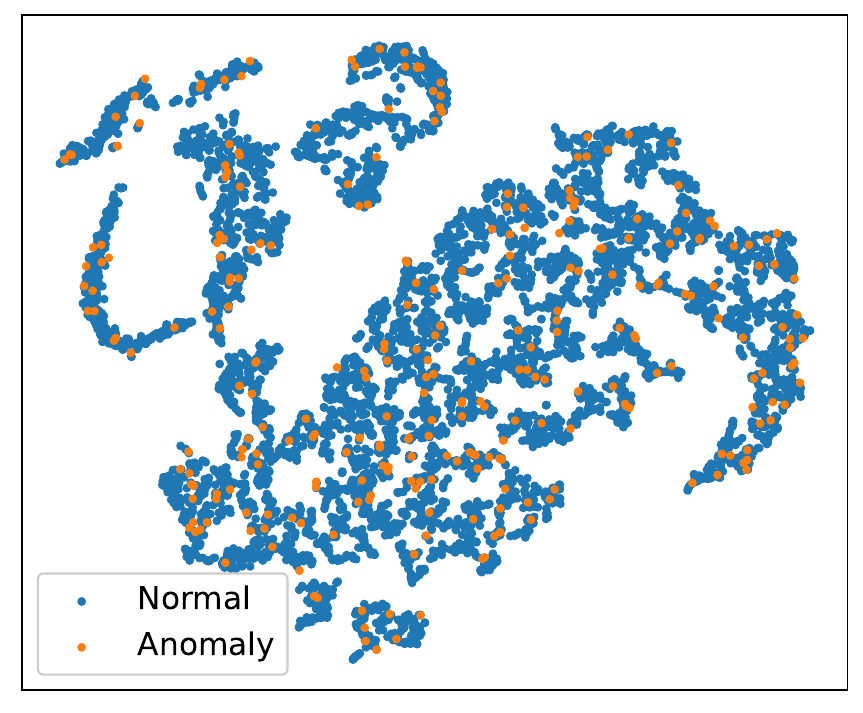}
    \caption{YelpHTL$\rightarrow$Amazon (meta-GDN)}
     \label{fig: amz-to-htl fewshot}
\end{subfigure}
\begin{subfigure}[b]{0.33\linewidth}
    \centering
    \includegraphics[height=4.5cm]{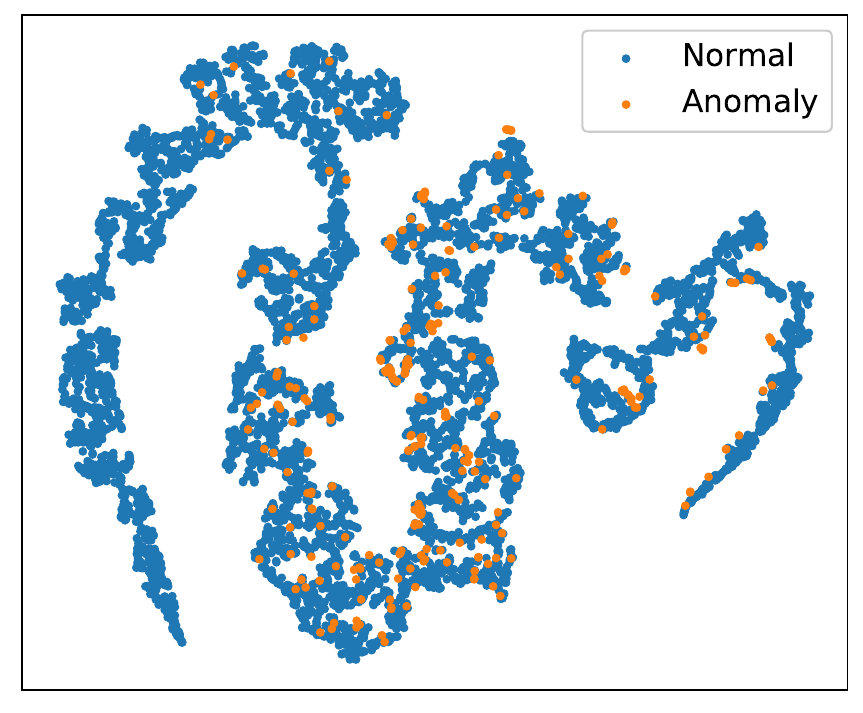} 
    \caption{YelpHTL$\rightarrow$Amazon (ACT)}
    \label{fig: htl-to-nyc fewshot}
\end{subfigure}
\begin{subfigure}[b]{0.33\linewidth}
    \centering
    \includegraphics[height=4.5cm]{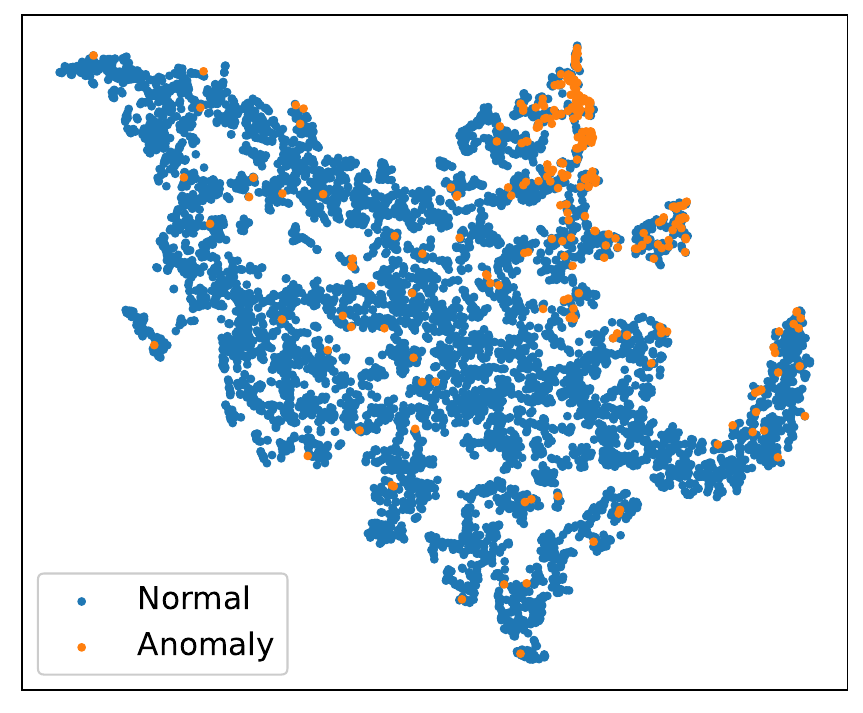}
    \caption{YelpHTL$\rightarrow$Amazon (CDFS-GAD)}
\end{subfigure}
\caption{t-SNE node embeddings of various GAD methods under the one-shot task. Each color denotes a different category.}
\label{fig: visualization}
\end{figure*}

\subsection{Parameters Sensitivity}

Figure~\ref{fig: sensitivity} presents the impact of hyper-parameters $\alpha$, $m$, $\beta_1$ and $\beta_2$ under one-shot task, as measured by AUC-ROC. The balance weight $\alpha$ controls the influence of the contrastive learning module, which includes the intra and inter-contrastive components. Figure~\ref{fig: balance_weight} shows that performance drops moderately at the two ends of $\alpha$. This is reasonable as a lower $\alpha$ may weaken the ability of our model to learn robust and domain-invariant representations, whereas a higher $\alpha$ risks overshadowing the primary loss function. For the prompt tuning module, the basis number $m$ exhibits a larger impact on the data pair YelpRes and Amazon compared to others. In particular, a larger basis is more beneficial for adapting to the substantial domain gap between the Yelp and Amazon datasets, as it grants the model increased flexibility, enabling more precise extraction of domain-specific knowledge. In terms of the self-training, we observe an optimal performance band when $\beta_1$ and $\beta_2$ are set around 0.03 and 0.25, respectively. Too few anomalies may not provide a strong enough learning signal, while too many may introduce noises that adversely affect the performance. Additionally, a higher $\beta_2$ reflects the real-world distribution of normal behavior, which typically outnumbers anomalies, thereby guiding the model to a more accurate representation of normality. Overall, despite small variations across different values of the hyper-parameters, the performance of CDFS-GAD still exhibits relative stability in terms of AUC-ROC. This suggests that our model is relatively less sensitive to various parameter settings.

\subsection{t-SNE Visualization}

Figure~\ref{fig: visualization} presents t-SNE visualizations of the node embeddings generated by three high-performing models in the one-shot setting. The embeddings from CDFS-GAD demonstrate a clear separation between normal and anomalous nodes, as evidenced by the distinct clusters of orange (anomalous) and blue (normal) points. In contrast, the visualizations for meta-GDN and ACT display a more interspersed arrangement, with anomalies mingling among normal samples. These findings further demonstrate the efficacy of CDFS-GAD in generating discriminative embeddings. This can be attributed to our model's effective utilization of anomaly information from the richly labeled source domain under the supervision of a scarcely labeled target domain. 

\section{Conclusion}

In this paper, we present the first investigation into cross-domain few-shot graph anomaly detection and also propose the CDFS-GAD framework to address this challenge. CDFS-GAD effectively bridges the gap between source and target domains through the integration of a domain-adaptive graph contrastive learning module and a prompt tuning module. Together, these components facilitate the extraction of common features across domains while preserving the unique, domain-specific features of each domain. To effectively utilize the few-shot labels from the target domain, we proposed a domain-adaptive hypersphere classification loss complemented by a self-training stage, both of which are crucial for optimizing performance in environments with scarce labeled data. These integrated components collectively enhance the detection accuracy in the target domain under few-shot constraints. Our experimental results confirm the effectiveness of each component and demonstrate superior performance across various domain pairs and few-shot conditions in comparison to existing GAD methods.

\section*{Acknowledgements}

This work was supported in part by the National Key Research and Development Program of China under Grant 2022YFF0712300, in part by the Fundamental Research Funds for the Central Universities under Grant YCJJ20241203.

\bibliographystyle{IEEEtran}
\bibliography{mybib}

\end{document}